\documentclass{article} %
\usepackage{iclr2027_conference,times}

\usepackage{amsmath,amsfonts,bm}

\def\eqref#1{equation~\ref{#1}}

\def\1{\bm{1}}

\DeclareMathAlphabet{\mathsfit}{\encodingdefault}{\sfdefault}{m}{sl}
\SetMathAlphabet{\mathsfit}{bold}{\encodingdefault}{\sfdefault}{bx}{n}

\newcommand{\cmark}{\textcolor{green}{\ding{52}}} %
\newcommand{\xmark}{\textcolor{red}{\ding{56}}} %
\newcommand{\tmark}{\textcolor{orange}{\ding{115}}} %

\usepackage{hyperref}

\usepackage[capitalise]{cleveref} %
\usepackage{pifont} %
\usepackage[dvipsnames]{xcolor} %
\usepackage{booktabs} %
\usepackage{comment} %
\usepackage{multirow} %
\usepackage{url}
\usepackage{graphicx}
\usepackage{amssymb}
\usepackage{lipsum}
\usepackage{makecell}

\title{UniState: Unifying Memory and Intervention for Controllable World Simulation}

\author{
Jaeseok Jeong$^{1,2}$ \quad
Kyungmook Choi$^{1}$ \quad
Sohyun Chung$^{1}$ \\
\textbf{Youngsik Yun$^{1}$ \quad
Youngjung Uh$^{1}$\thanks{Corresponding author}} \\[0.5em]
$^{1}$Department of Artificial Intelligence, Yonsei University \quad
$^{2}$Familiar\\[0.5em]
{\small\texttt{\{jete\_jeong,kyungmook.choi,sohyun.chung,}}\\
{\small\texttt{bbangsik,yj.uh\}@yonsei.ac.kr}} \\ [0.5em]
\small\textbf{Project Page:} \url{https://curryjung.github.io/unistate/} %
}

\iclrfinalcopy %
\begin{document}
\maketitle
\lhead{Preprint}
\begin{abstract}
World models offer increasingly rich environments for exploration, yet intervening in a world while observing the consequences from freely chosen viewpoints remains challenging. We tackle these challenges by introducing \textbf{UniState}, a framework for controllable world simulation built on a unified world-state representation. This representation unifies spatial memory and user interventions. Users specify object motion directly within spatial memory, while independently choosing the camera trajectory from which the resulting world state is observed. This unification makes the relationship between the scene and the requested changes explicit, removing ambiguity between object motion and viewpoint change from the control specification. Crucially, specifying an intervention does not require prescribing every resulting movement: the video model can synthesize additional motion within controlled objects and plausible interactions with their surroundings. We develop training and sampling strategies that help generated videos follow the specified object motion and camera trajectory while preserving this capacity to generate responses beyond the prescribed transformations. Experiments demonstrate more accurate joint object and camera control than the existing methods, alongside qualitative evidence that intervention-induced scene changes persist across rollouts.

\end{abstract}
\begin{figure}[t]
\centering
\includegraphics[width=\linewidth]{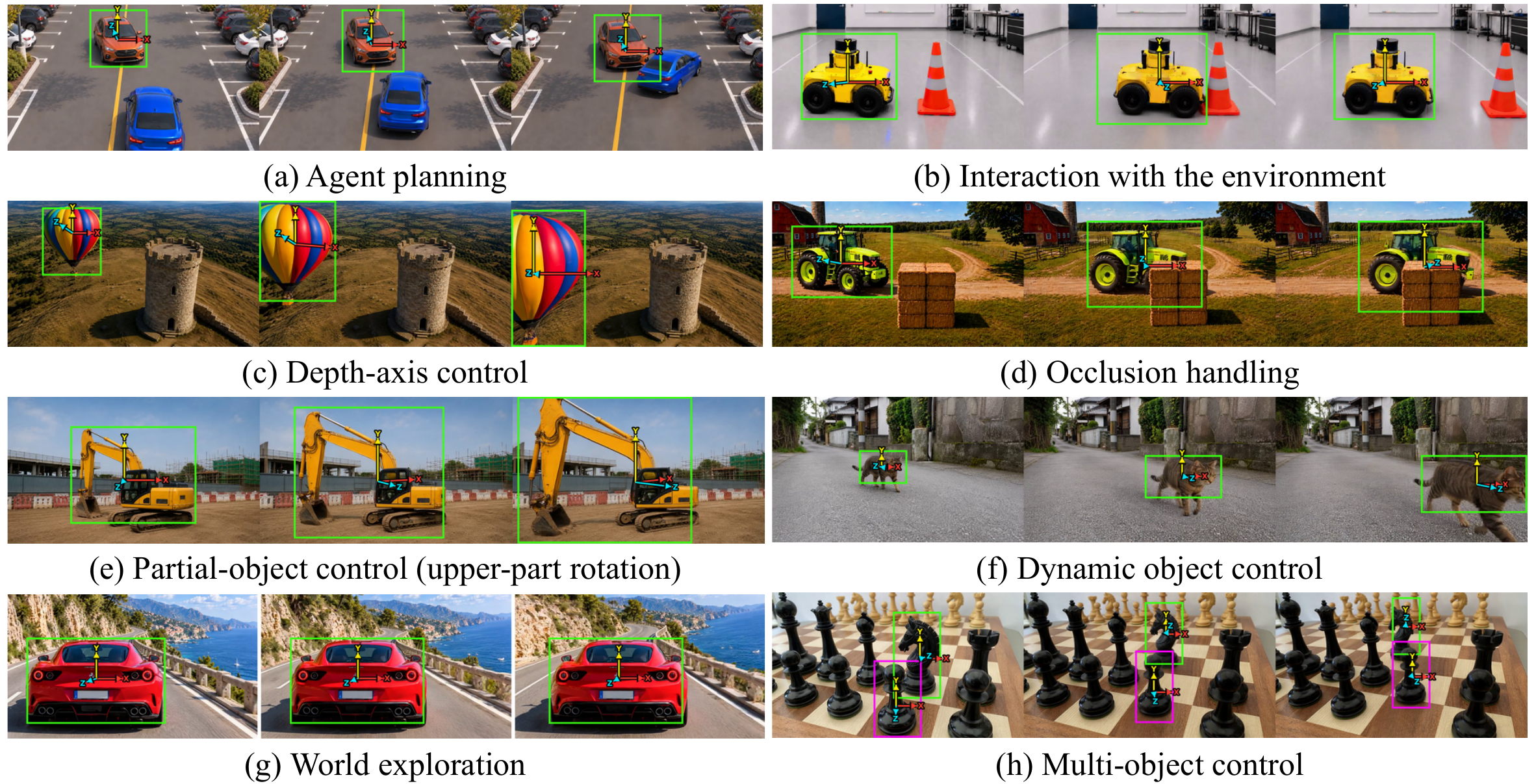}
\caption{\textbf{UniState} is a framework for controllable world simulation, enableing coherent and interactive world simulation.}
\label{fig:gap-patch-teaser}
\end{figure}

\section{Introduction}
\label{sec:introduction}
World simulation models have been built from video models for their increasing performance of producing high-fidelity future predictions given an initial frame and a description of the events occurring in the scene~\citep{wan2025,nvidia2026cosmos3}. They can be used as explorable environments for training agents, especially with camera controls~\citep{wang2025ati,xing2025motioncanvas} and object controls~\citep{wang2024boximator,geng2024motionprompting}.

However, using world models for learning agents is limited to viewpoint exploration~\citep{bar2024navigationworldmodels} even though recent world models cover complex real-world environments, and training agents with \textit{physical intervention} to the world is limited to restricted configurations such as grippers with a fixed camera~\citep{du2023unipi,yang2023unisim,huang2025vid2world}. We attribute this gap to recent world models that cover navigation and omit physical intervention.

We suggest that physical intervention beyond navigation is a crucial component for training physical agents, as well as other important requirements including spatio-temporal coherence and free-viewpoint observation. Existing video models exploit persistent spatial memory to support spatio-temporal coherence~\citep{xiao2025worldmem,yu2026mosaicmem}, camera control to support free-viewpoint observation~\citep{he2024cameractrl,he2025cameractrlii}, and dedicated conditions to support object motion~\citep{wu2024draganything,wang2025ati}.
However, supporting physical intervention in a persistent world with free-viewpoint observation remains challenging.

We introduce UniState, capable of above three requirements. Briefly, it contains initial spatial memory and user intervention in the same world coordinate system in order to work as a spatio-temporal twin of the world and allow the viewpoint to be separately specified. It removes the ambiguity that the same trajectories can be achieved by either object motion or viewpoint control in previous methods.
The initial spatial memory is the latents lifted into 3D from an initial image~\citep{yu2026mosaicmem}. Then users specify motion of a region with a sequence of 6-DoF transformations, enabling 3D object translation and rotation, as illustrated in \cref{fig:gap-patch-teaser}.
Finally, a camera trajectory determines the viewpoints from which UniState is rendered. Last but not least, we introduce sampling and training techniques to reflect all the above variables.

This unified state guides video generation while preserving the model's capacity to synthesize plausible responses to the specified interventions. For example, an intervention in the surrounding environment can lead the model to generate an agent's evasive maneuver, such as taking a detour to avoid an approaching vehicle (\cref{fig:gap-patch-teaser}(a)). Controlling a bulldozer can similarly induce motion in nearby objects through contact, without explicitly prescribing their trajectories (\cref{fig:gap-patch-teaser}(b)).
This flexibility also extends to motion within the controlled object: specifying only the translation of a cat's torso allows the model to generate the accompanying leg movements for walking (\cref{fig:gap-patch-teaser}(f)).

 To guarantee that the video model to faithfully translate the unified representation into observations from the specified viewpoint, we design both sampling and training strategies for coherent joint object and camera control. Our sampling strategy promotes consistent generation from the specified viewpoint during object interventions, while our training strategy reduces the mismatch between training and inference conditions, helping the video model follow the specified object motion and camera trajectory while retaining its capacity to generate plausible responses.

Experiments show that UniState achieves more accurate joint object and camera control than the evaluated baselines.
Qualitative results demonstrate manipulation of whole objects and selected parts with consistent depth ordering and occlusion, alongside motion and interactions beyond the prescribed transformations.
Ablation studies support the benefits of the proposed training and sampling strategies for control accuracy.

Our key contributions are:
\begin{itemize}
    \item With UniState, we show that unifying spatial memory and
    user interventions in a unified world-state representation
    allows object motion and camera trajectory to be specified
    independently.

    \item We show that using the initial state as a reference during
    sampling and reducing the mismatch between training and inference
    conditions improve joint object and camera control.
\end{itemize}

\begin{table}[t]
    \centering
    \caption{
    Comparison of spatial memory and object-intervention representations. UniState is the only method that unifies spatial memory and user intervention within a unified representation and world coordinate system. Detailed capability and intervention-representation comparisons are provided in
    \cref{tab:capability_comparison} and \cref{tab:intervention_representation}.
    }
    \label{tab:representation_comparison}
    \begin{tabular}{lccc}
    \toprule
    Method
    & Spatial memory
    & User intervention
    & \makecell{Unified repr. \\} \\
    \midrule
    VerseCrafter
        & RGB point cloud
        & 3D Gaussians
        & \xmark \\
    WorldDirector
        & Video context tokens
        & 2D box trajectories
        & \xmark \\
    MotionCanvas
        & Image latent
        & 2D boxes / point tracks
        & \xmark \\
    SymphoMotion
        & Point-cloud render
        & 3D point trajectories
        & \xmark \\
    Real2SAM2Real
        & Video latent
        & 3D mesh
        & \xmark \\
    Ours
        & \multicolumn{2}{c}{UniState}
        & \cmark \\
    \bottomrule
    \end{tabular}

\end{table}

\section{Related work}
\label{sec:relwork}

\paragraph{Video world models as interactive simulators.}
Video world models simulate the consequences of actions for planning and policy learning, using generated videos as plans for control~\citep{du2023unipi} or as interactive environments for policy training~\citep{yang2023unisim}.
Action-conditioned video models further leverage pretrained video priors or learn from latent actions and recorded controls for interactive simulation~\citep{rigter2024avid,huang2025vid2world,bruce2024genie,alonso2024diamond,valevski2024gamengen,li2025hunyuangamecraft,he2025matrixgame2}.
Recent methods extend exploration across diverse scenes, using context management and accelerated generation to sustain interaction~\citep{mao2025yume15,sun2025worldplay,robbyantteam2026lingbotworld,parkerholder2025genie3}.
Complementary approaches use predicted observations for navigation planning~\citep{bar2024navigationworldmodels}, incorporate geometric prediction~\citep{aetherteam2025aether,chen2025deepverse}, or ground exploration in geographically indexed street views~\citep{seo2026seoulworldmodel}.
Instruction-driven behaviors and object-directed actions further broaden interaction beyond navigation~\citep{tang2025hunyuangamecraft2,gao2026lingbotworld2}.
UniState extends world exploration with time-varying 6-DoF interventions on selected object regions with the unified representation and world coordinate system as persistent memory, while allowing the camera trajectory to be specified independently.

\paragraph{Object control and camera control.}
Methods for controlling motion in video generation specify motion through point trajectories~\citep{wu2024draganything,zhang2024tora,geng2024motionprompting,wang2025ati} or bounding boxes~\citep{wang2024boximator}; joint object and camera control further requires coordinating these motion signals with viewpoint conditions~\citep{wang2023motionctrl,xing2025motioncanvas}, often using depth, pose, or 3D guidance~\citep{wang2024levitor,ji2025posetraj,gu2025diffusionasshader,cao2025uni3c,wang2026worlddirector,zhang2026symphomotion}.
Since viewpoint change and object motion both induce image-space displacement, recent methods disentangle them through cross-view transfer~\citep{liu2026moright} or dynamic reprojection of world-anchored trajectories~\citep{gu2026worldcraft}.
Methods using explicit 3D representations condition video generation on Gaussians expressed in a shared coordinate frame~\citep{zheng2026versecrafter} or editable foreground caches~\citep{wu2026real2sam2real}, while appearance-aware methods transport latent features along point tracks~\citep{chu2025wanmove,gu2026worldcraft}.
As summarized in \cref{tab:representation_comparison}, the compared methods encode spatial memory and user intervention in separate representations.
This separation leaves generation susceptible to background drift and ambiguity between object motion and viewpoint change.
UniState grounds both in a shared appearance-bearing 3D representation and world coordinate system, making their spatial relationship explicit.

\paragraph{Persistent and editable spatial memory.}
While camera conditioning controls the viewpoint~\citep{he2024cameractrl,he2025cameractrlii}, scene consistency can be maintained by retrieving spatially relevant observations~\citep{xiao2025worldmem,yu2025contextasmemory,sun2025worldplay} or compressing historical context~\citep{zhang2025framepack,oshima2025worldpack,hong2025relic}.
Retrieval and compression retain scene evidence but do not themselves define geometric operations on entities.
Geometry-based approaches render, warp, or retrieve scene information using explicit spatial memories~\citep{ren2025gen3c,wang2026latentspatialmemory,zhao2025spatia,li2025vmem}, with some excluding dynamic objects from persistent memory~\citep{wang2026latentspatialmemory}.
Other memory-based approaches maintain out-of-view dynamics~\citep{chen2026hydra,duan2026liveworld} or combine coordinated 3D object and camera trajectories with persistent dynamic object memory~\citep{wang2026worlddirector}.
Spatial patch memories further support relocation, duplication, and deletion, but focus on scene composition rather than prescribed instance-level motion over time~\citep{yu2026mosaicmem}.
Explicit spatial representations also support editing through evolving voxel states, but validation of this approach remains limited to game environments~\citep{garcin2026persist}.
UniState applies time-varying 6-DoF transformations to whole or partial object regions within shared 3D spatial memory, while the video prior synthesizes dynamics beyond the prescribed rigid motion.
\begin{figure}[t]
\centering
\includegraphics[width=\linewidth]{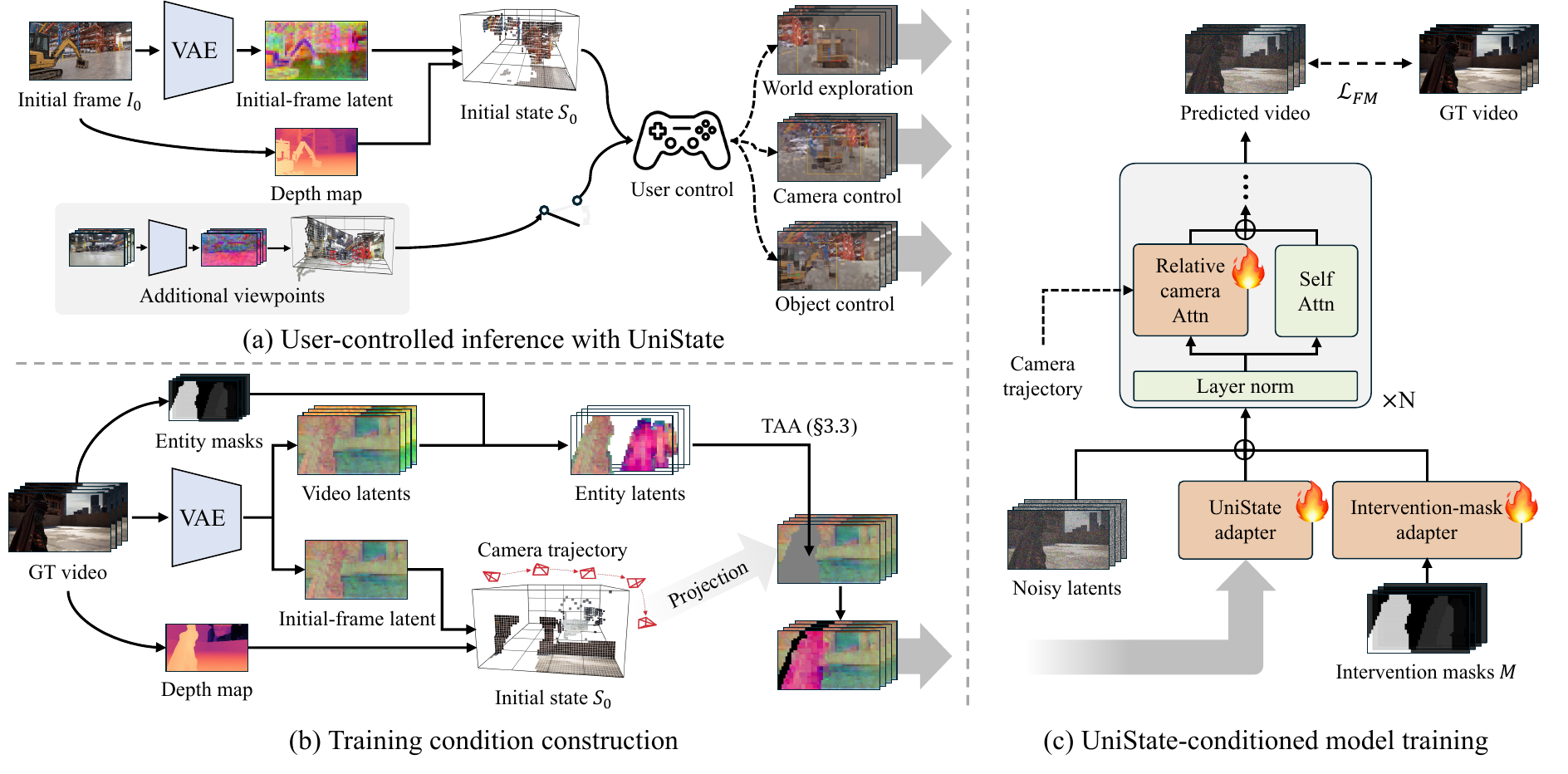}
\caption{\textbf{Overview of UniState.}
(a) Source observations are lifted into a unified state.
Users specify interventions and a camera trajectory to construct the inference conditions.
(b) During training, entity latents from ground-truth videos are combined with initial-state renders by masked compositing.
(c) %
Only the conditioning modules marked with flames are trained.
$\mathcal{L}_{\mathrm{FM}}$ denotes the flow-matching loss between predicted and target latent velocities.
}
\label{fig:overview}
\end{figure}

\section{Method}
\label{sec:method}

\textbf{UniState} generates a video from an initial frame, user-specified interventions, and a camera trajectory.
We lift initial-frame latent features into 3D to construct spatial memory, then express user interventions by transforming the geometry of selected regions within this memory.
The transformed entities and unchanged scene context thus remain in a single set of 3D latents, which we call a unified \textit{world-state representation}.
Rendering this state along an independently specified camera trajectory provides the conditions for video generation.
\cref{fig:overview} summarizes the inference and training pipelines.

We first describe how UniState constructs, manipulates, and renders this unified state (\cref{sec:unistate}).
We then introduce Scene-Anchored Joint Guidance (\cref{sec:sajg}) and a training strategy (\cref{sec:training}) that help the video model follow the resulting conditions.

\subsection{Unified World-State Representation}
\label{sec:unistate}
\paragraph{Constructing spatial memory.}
We encode the initial image into VAE latent features and lift them into 3D using the source camera and estimated depth~\citep{wang2026moge}. Each resulting 3D latent pair has geometry with its source feature; together, these latents form the initial state $S_0$. Additional views of the same initial scene can extend $S_0$ by lifting and registering their latent features in the same world coordinate system. Geometric construction details are provided in Appendix~\ref{asec:gap_patch}.

\paragraph{Expressing interventions within the state.}

Source-view segmentation masks, provided by the user or obtained
with SAM~2~\citep{ravi2025sam2}, identify the latent index set
$\mathcal{I}_k$ for each entity $k$, which may be a whole object
or a selected part.
For each entity, the user specifies rigid\footnote{Part-wise rigid
transformations can approximate non-rigid motion.}
transformations $\{G_{t,k}\in\mathrm{SE}(3)\}_{t=1}^{T}$
in the shared world coordinate system over $T$ subsequent frames.
Let $X_{0,i}$ denote the geometry of the $i$-th 3D latent
in the initial state $S_0$. At each time $t$, its geometry is
\begin{equation}
X_{t,i}=
\begin{cases}
G_{t,k}X_{0,i}, & i\in\mathcal{I}_k,\\
X_{0,i}, & \text{otherwise}.
\end{cases}
\label{eq:entity_transform}
\end{equation}
We construct each intervened state $S_t$ directly from $S_0$
using these transformed geometries, while preserving source
latent features and leaving unselected regions unchanged.

\paragraph{Rendering the state along the camera trajectory.}
Given an independently specified camera trajectory
$C=\{C_t\}_{t=1}^{T}$, a visibility-aware renderer $\mathcal{R}$
projects the intervened state into each camera view: $(W_t,M_t)=\mathcal{R}(S_t;C_t).$
Here, $W_t$ contains the projected source latent features, and $M_t$ marks the visible regions of selected entities.
Visibility is resolved against the complete scene; rendering details are provided in Appendix~\ref{asec:gap_patch}.

The UniState adapter and intervention-mask adapter independently project the patchified latent condition $W=\{W_t\}_{t=1}^{T}$ and intervention masks $M=\{M_t\}_{t=1}^{T}$ into the video-token dimension. Their outputs are added to the input video tokens once, before the transformer blocks.
The camera-conditioning module instead operates in every transformer block: a parallel attention branch uses the relative-ray camera representation and adds its output to the backbone self-attention output.
These conditions guide generation while the video model synthesizes the remaining scene content and motion.

\paragraph{Interaction and state updates.}
User interventions specify the motion of the controlled entity, while the video model synthesizes plausible responses in the surrounding scene.
Conditioning scene regions involved in an interaction on their previous configuration can constrain these responses.
We therefore temporarily exclude these regions from the memory
condition while retaining the prescribed control over the entity.
This allows nearby objects to move in response to contact without explicitly prescribing their trajectories.

After the interaction, we obtain updated geometry and latent features for the affected regions from generated observations, register the geometry to the shared world coordinate system, and update the corresponding state regions.
The updated state serves as $S_0$ for the next rollout, allowing subsequent viewpoint changes and interventions
to build on the resulting scene configuration.

\subsection{Scene-Anchored Joint Guidance}
\label{sec:sajg}
Our conditional video model $f_\theta$ is trained with a flow-matching objective and predicts latent velocities during sampling.

Although the unified state specifies interventions independently of camera viewpoint, the generated video may still entangle their effects, causing background motion to deviate from the prescribed camera trajectory. Our key observation is that this ambiguity can be removed by constructing a camera-only reference: if we render the initial state $S_0$ under the same camera trajectory but without applying any intervention, the resulting condition should contain scene changes induced by viewpoint change alone. 
We therefore use the velocity conditioned on the intervention for the intervened region and use the velocity conditioned on the initial state for the background. 
We spatially route between these two predictions and blend the result with the full joint prediction. We call this procedure Scene-Anchored Joint Guidance (SAJG).

\paragraph{Scene-anchor and joint predictions.}
Using the same camera trajectory, we render the initial state $S_0$
without interventions to obtain the scene-anchor condition
$W_s$, and the intervened states $\{S_t\}_{t=1}^{T}$ to obtain
the joint condition $W_j$, corresponding to $W$ in \cref{sec:unistate}.

Using the same noisy video latent, denoising timestep, camera
condition, source observation, and text prompt, we evaluate
$f_\theta$ under the scene-anchor and joint conditions to obtain
the respective velocity predictions $V_s$ and $V_j$.
Both predictions use classifier-free guidance; implementation
details are provided in Appendix~\ref{app:sajg}.

\paragraph{Spatial routing.}
The scene-anchor prediction provides a reference for
camera-induced changes in the surrounding scene, while the
joint prediction incorporates the requested interventions.
We therefore use the joint prediction in regions affected by
these transformations, including the entities' target locations
and the regions they vacate.
A soft scene mask $G_s$ excludes these projected source and
target regions, combining the predictions as
\begin{equation}
V_{\mathrm{routed}}
=
G_s\odot V_s+(1-G_s)\odot V_j,
\label{eq:sajg_masked}
\end{equation}
where $\odot$ denotes element-wise multiplication.
Mask construction details are provided in Appendix~\ref{app:sajg}.

\paragraph{Joint prediction anchoring.}
To retain the joint prediction's contribution throughout the video,
including scene regions assigned to the scene-anchor prediction,
we blend the spatially routed prediction with the full joint prediction:
\begin{equation}
V_{\mathrm{final}}
=
\alpha V_j+(1-\alpha)V_{\mathrm{routed}},
\qquad \alpha\in[0,1].
\label{eq:sajg_final}
\end{equation}
Here, $\alpha$ controls the weight of the full joint prediction
in the final blend, ensuring a minimum joint contribution of
$\alpha$ at every spatio-temporal location.
Setting $\alpha=0$ uses the spatially routed prediction alone,
whereas $\alpha=1$ uses only the joint prediction.
We use $\alpha=0.25$ unless stated otherwise.
At each denoising step, $V_{\mathrm{final}}$ is passed to the
flow-matching solver.

\subsection{Training}
\label{sec:training}

\paragraph{Training conditions.}
We train on videos paired with camera conditions $C$, latent
conditions $W$, and intervention masks $M$ derived from per-frame
object masks.
We construct $W$ by combining initial-state renders with
ground-truth entity latents.
The latter preserve alignment with the target video, avoiding
misalignment and floating artifacts caused by transporting source
features with imperfect geometry
(see \cref{fig:source-transport-visualization} in the appendix).

We render the initial state from each target camera and zero out the selected entities' projected source regions. We then insert ground-truth entity latents at their target locations, retaining the initial-state render elsewhere. Vacated source regions remain zero unless covered by a target entity. The exact compositing formulation is provided in Appendix~\ref{app:training}.

\paragraph{Transport-Aligned Augmentation.}
The composited conditions introduce a training--inference mismatch:
training uses entity latents from each ground-truth target frame,
whereas inference transports source latents through the transformed
geometry.
To reduce this mismatch, we apply Transport-Aligned Augmentation
(TAA), which spatially coarsens future-frame entity latents while
preserving their target support.
TAA modifies only the latent condition; the intervention masks and
ground-truth training targets remain unchanged.
The augmentation procedure is detailed in
Appendix~\ref{app:training}.

\paragraph{Curriculum training.}
We train UniState in two stages using a flow-matching objective \(\mathcal{L}_{\mathrm{FM}}\),
keeping the pretrained video backbone, VAE, and text encoder frozen.
\citep{ling2024dl3dv}.
The first stage trains the camera-conditioning module and UniState adapter on DL3DV \citep{ling2024dl3dv}.

Starting from these weights, the second stage adds the intervention-mask adapter and jointly trains all three modules on HM-World\citep{chen2026hydra} using the composited conditions and TAA described above.

\section{Experiments}
\label{sec:exp}

\begin{figure}[t]
\centering
\includegraphics[width=0.85\linewidth]{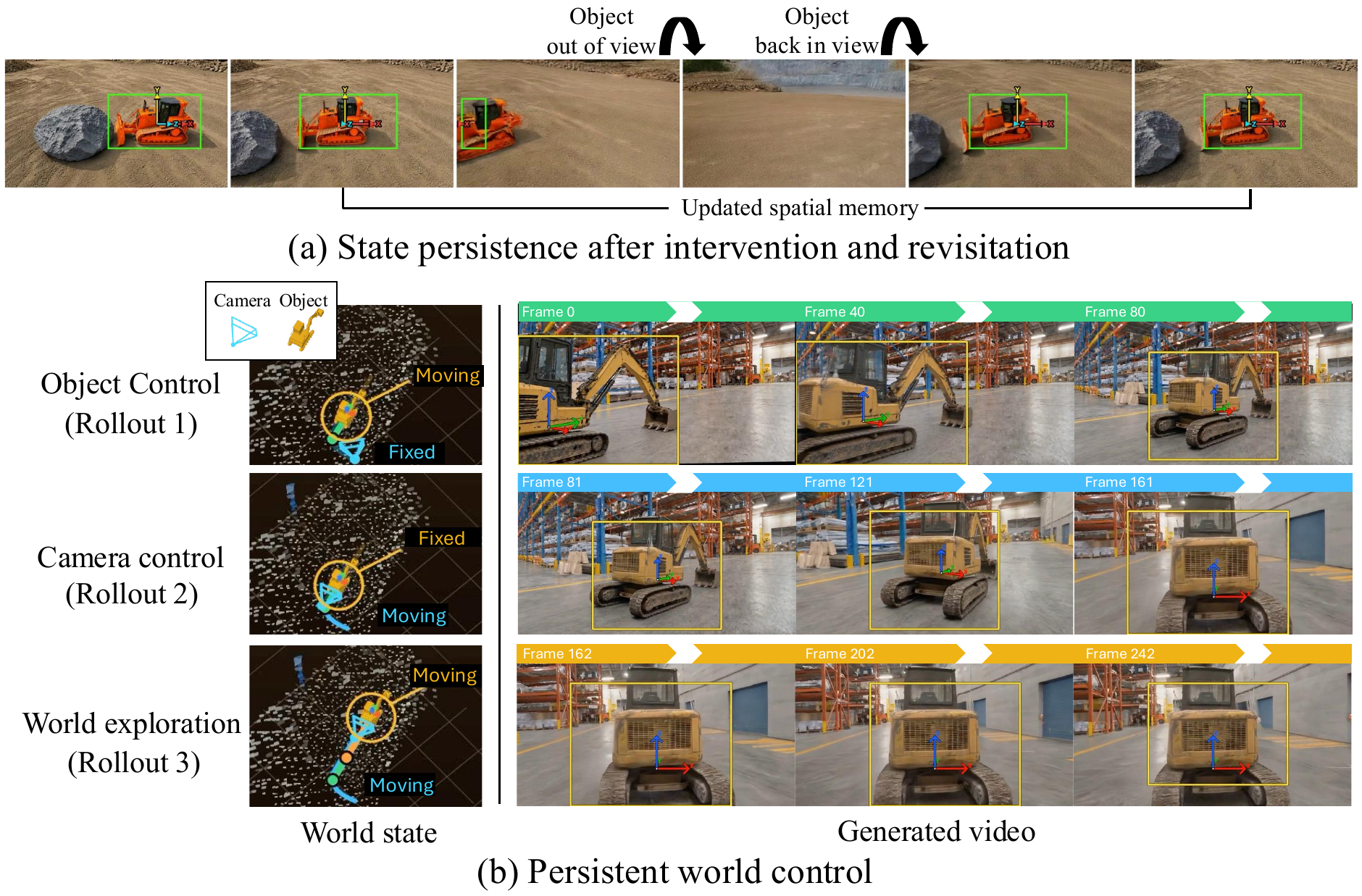}
\caption{\textbf{State persistence and control across rollouts.}
(a) A rock displaced by the controlled bulldozer remains in its updated position after the camera leaves and revisits the region.
(b) Three successive rollouts demonstrate object control with a fixed camera, camera control with a fixed object pose, and joint object and camera control, while carrying the world state forward.
The left column visualizes the unified state and prescribed controls; the right columns show generated frames.
Colored boxes identify controlled entities, and coordinate axes indicate their prescribed poses.}
\label{fig:interaction-revisit}
\label{fig:ours-qualitative}
\end{figure}

\subsection{Implementation details}
We use a pretrained Cosmos3-Nano I2V model~\citep{nvidia2026cosmos3} and follow the two-stage curriculum in \cref{sec:training}, with 20K optimization steps on DL3DV followed by 10K on HM-World.
We generate 81-frame videos at $832\times480$ resolution and 16\,fps.
Optimization settings and caption construction are detailed in Appendix~\ref{app:training}.

\subsection{Evaluation} 
We evaluate all methods on the same 277 scenarios spanning
diverse camera trajectories and user interventions.
Each method generates two videos per scenario with independent
random seeds (554 videos total), using an identical evaluation
protocol and averaging metrics across the two seeds.

\paragraph{Joint control accuracy}
We introduce World-Control Projection Error (WCPE) to evaluate object control specified in world coordinates. Because video models produce 2D observations, an object's screen-space position alone does not establish whether it follows the specified motion in the world. We therefore assess object motion relative to the surrounding scene, while checking that the scene remains aligned with the specified camera trajectory. WCPE implements this assessment as a 2D proxy, using target projections to evaluate object–background relative positions and background alignment. Requiring both prevents correct screen-space object placement alone from yielding a low error when the surrounding scene is misaligned. The metric formulation and evaluation protocol are detailed in \cref{app:wcpe}.

\paragraph{Camera-control accuracy} We evaluate camera-control accuracy by comparing the camera trajectory recovered from each generated video with its target trajectory. We report Cam-T, the mean translation error between the recovered and target camera positions, and Cam-R, the mean geodesic rotation error between their orientations, measured in degrees. Lower values for both metrics indicate more faithful camera control.

\subsection{Qualitative Results}
\label{sec:qualitative-results}
\paragraph{Diverse control through user interventions.}
\cref{fig:gap-patch-teaser}(c) and (d) demonstrate depth-axis control and occlusion handling consistent with scene geometry.
Interventions can also target selected object parts: in \cref{fig:gap-patch-teaser}(e), the user rotates the excavator's upper structure independently of its tracks, specifying the rotation direction and angular speed.
For non-rigid objects, partial control allows the video model to complete unspecified articulation.
In \cref{fig:gap-patch-teaser}(f), the user specifies only the cat's torso trajectory, while the model generates accompanying
limb motion and maintains a coherent body configuration.
\cref{fig:gap-patch-teaser}(g) and (h) further demonstrate camera-controlled world exploration and simultaneous control of multiple objects, respectively.

\paragraph{State persistence and control across rollouts.}
In \cref{fig:interaction-revisit}, the controlled bulldozer displaces a passive rock.
After the resulting state is incorporated into memory, the camera moves away until the affected region is out of view, then returns.
The rock retains its displaced position upon revisitation, providing qualitative evidence of state persistence across rollouts.
\cref{fig:interaction-revisit}(b) demonstrates continued control of the same world across three successive rollouts.
The user first moves the excavator with the camera fixed, then changes the viewpoint while preserving the excavator's updated world-space pose, and finally controls both the object and camera.
Each rollout continues from the preceding state, allowing the user to manipulate an entity, observe the resulting configuration from another viewpoint, and intervene again.
This separates control of the entity's world state from control of the observation viewpoint.

\subsection{Ablation Study}
\label{sec:ablation}
\begin{figure}[t]
\centering
\includegraphics[width=0.80\linewidth]{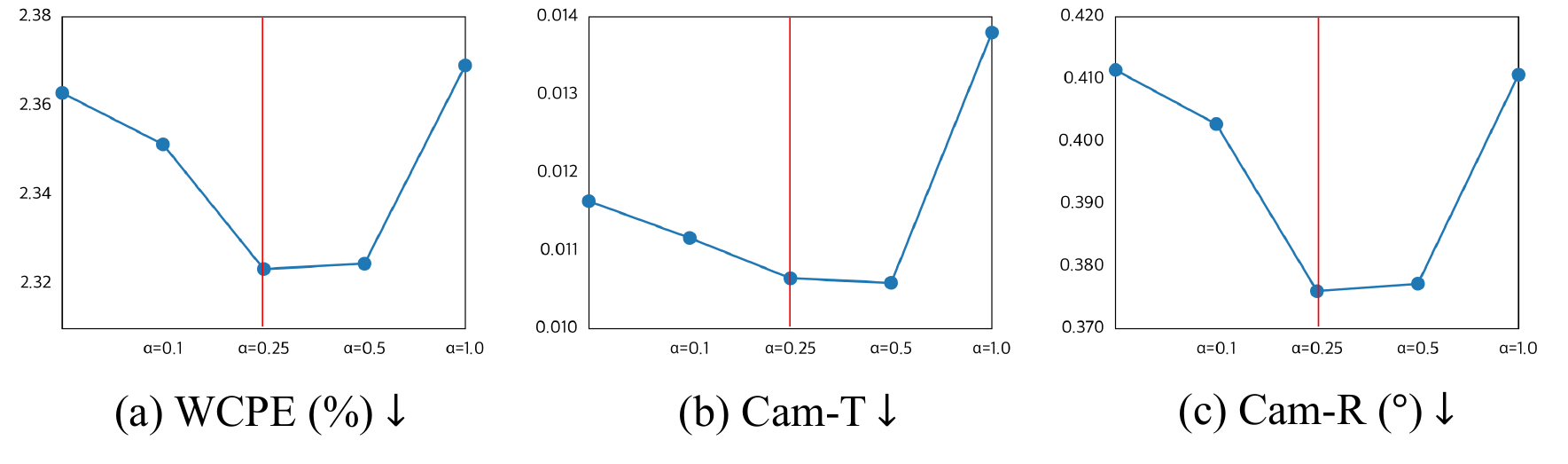}
\caption{Ablation results for Scene-Anchored Joint Guidance.
The red vertical line denotes the selected setting \(\alpha=0.25\).
}
\label{fig:sajg-ablation}
\end{figure}

\paragraph{Scene-Anchored Joint Guidance.}
\cref{fig:sajg-ablation} shows that intermediate values of $\alpha$ outperform both spatial routing alone ($\alpha=0$) and the joint prediction alone ($\alpha=1$).
We select $\alpha=0.25$, which achieves the lowest WCPE and Cam-R and nearly the lowest Cam-T.
These results support combining a scene-anchor correction with a global joint contribution.

\paragraph{Transport-Aligned Augmentation.}
We compare three ways of constructing entity latent conditions during training: transporting initial-frame latents using estimated geometry, using aligned ground-truth entity latents from each target frame, and applying TAA to those ground-truth latents.

All variants share training data, initialization, optimization budgets, inference conditions, and SAJG settings.
As shown in \cref{tab:ablation-taa}, replacing transported initial-frame latents with ground-truth latents reduces WCPE and Cam-T, but slightly increases Cam-R.
Applying TAA further reduces all three errors, supporting the combination of spatially aligned training conditions and augmentation that reduces their latent mismatch with inference-time transport.

\begin{table}[t]
\centering
\caption{Ablation study on Transport-Aligned Augmentation (TAA).
$^\dagger$ denotes using transported initial-frame latents as the entity condition during training.}
\label{tab:ablation-taa}
\begin{tabular}{lccc}
\toprule
Training condition & WCPE (\%) $\downarrow$ & Cam-T $\downarrow$ & Cam-R ($^\circ$) $\downarrow$ \\
\midrule
Direct source-transport$^\dagger$ & 2.857 & 0.0115 & 0.405 \\
GT condition w/o TAA & 2.574 & 0.0114 & 0.412 \\
GT condition + TAA & \textbf{2.323} & \textbf{0.0106} & \textbf{0.376} \\
\bottomrule
\end{tabular}
\end{table}

\paragraph{Curriculum training.}
Staged training outperforms interleaved training under matched data exposure and computation; details are provided in Appendix~\ref{app:curriculum-ablation}.

\subsection{Comparisons}
\label{sec:comparisons}
\begin{figure}[t]
\centering
\includegraphics[width=0.95\linewidth]{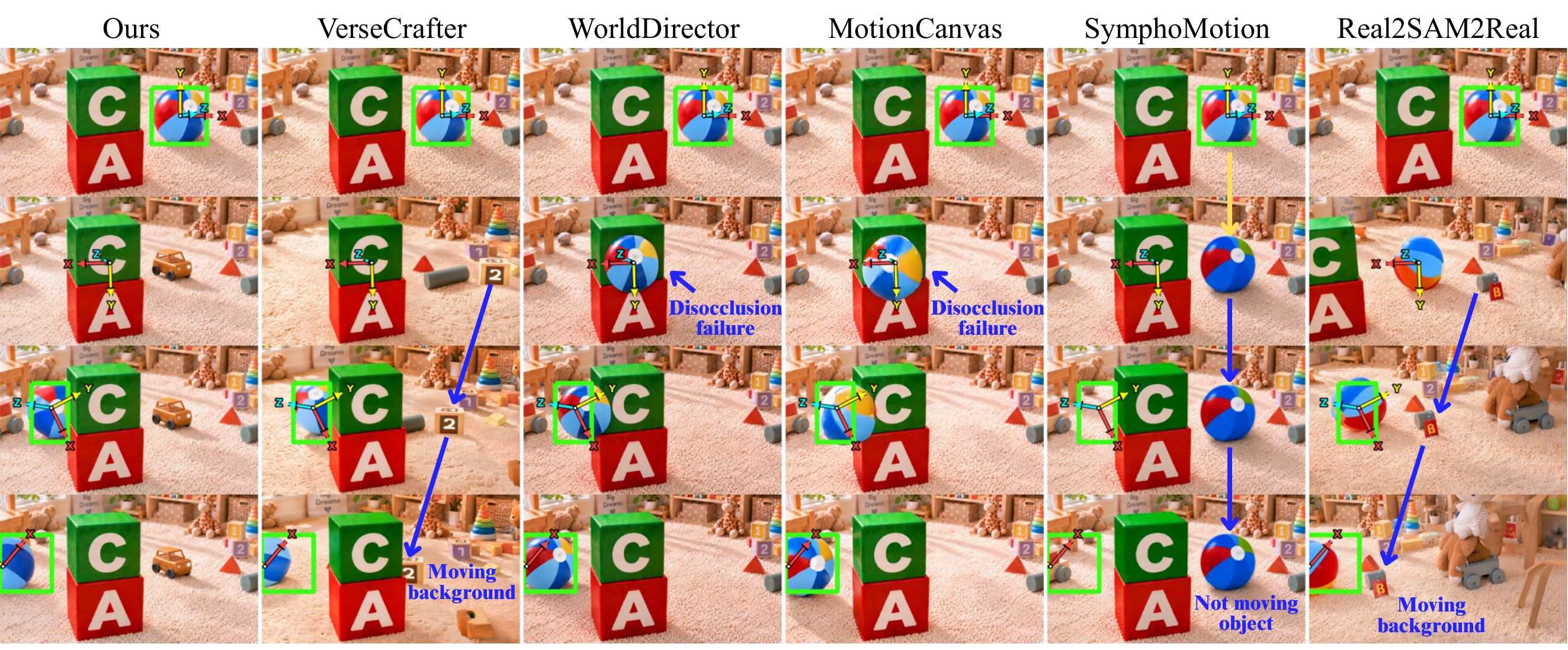}
\caption{\textbf{Object translation and rotation control under occlusion.}
The ball translates from right to left and rotates behind the
alphabet blocks. Green boxes and XYZ axes indicate its target
position and orientation. UniState preserves object motion,
occlusion, and background stability.}
\label{fig:vs-competitors-blocksandball}
\end{figure}

We compare our method with VerseCrafter~\citep{zheng2026versecrafter}, WorldDirector~\citep{wang2026worlddirector}, MotionCanvas~\citep{xing2025motioncanvas}, SymphoMotion~\citep{zhang2026symphomotion}, and 
Real2SAM2Real~\citep{wu2026real2sam2real}. 
We evaluate joint object and camera control using WCPE, Cam-T, and Cam-R. As shown in \cref{tab:comparison-competitors}, our method achieves the lowest WCPE, Cam-T, and Cam-R, demonstrating accurate joint control of object motion and camera viewpoint.
Qualitative comparisons in \cref{fig:vs-competitors-blocksandball} further show these failure modes. In this example, the user keeps the camera static while translating and rotating the ball laterally behind the alphabet blocks. VerseCrafter and Real2SAM2Real often absorb object motion into viewpoint change, resulting in undesired viewpoint change despite the specified static camera condition. This makes the ball appear to move with the camera rather than independently in the scene. WorldDirector and MotionCanvas preserve the static camera more reliably, but fail to reproduce the correct depth ordering and rotation of the ball. In contrast, our method faithfully follows the intended joint object and camera control, capturing object translation and rotation under a static camera while handling occlusion. See \cref{sec:more-comparisons} for additional qualitative comparisons. An additional comparison with MosaicMem is provided in \cref{app:comparison-mosaicmem}.

\begin{table}[t]
\caption{Comparison with competing methods on joint object and camera control.
\textbf{Bold} denotes the best score, and \underline{underline} denotes the second-best score for each metric.}
\label{tab:comparison-competitors}
\begin{center}
\begin{tabular}{lccc}
\toprule
Method & WCPE (\%) $\downarrow$ & Cam-T $\downarrow$ & Cam-R ($^\circ$) $\downarrow$ \\ 
\midrule
VerseCrafter            & 3.039          & 0.0540          & \underline{0.385}          \\
WorldDirector           & 2.487          & 0.0844  & 0.571 \\
SymphoMotion            & 2.822          & 0.0627          & 0.526          \\
MotionCanvas     & \underline{2.425}          & \underline{0.0378}          & 0.428          \\
Real2SAM2Real   & 3.061    & 0.104       &   0.608  \\
Ours                    & \textbf{2.323} & \textbf{0.0106} & \textbf{0.376} \\
\bottomrule
\end{tabular}
\end{center}
\end{table}

\section{Conclusion}
\label{sec:conclusion}
UniState unifies spatial memory and user interventions in a shared world coordinate system for controllable world simulation. This unified world-state representation allows object motion and camera trajectory to be specified independently. It provides spatial constraints for video generation while preserving the video model's capacity to synthesize plausible responses, so users can specify intended changes without prescribing every resulting motion or interaction. Quantitative comparisons and ablations support our training and sampling strategies for joint object and camera control across diverse camera trajectories and user interventions. Revisitation results further show that generated interaction outcomes can be incorporated into spatial memory and retained across rollouts. Together, these findings support physical intervention in a persistent world with free-viewpoint observation.

\paragraph{Limitations}
Interaction effects rely on the video model's generative prior: although controlled object states can be updated outside the camera view, their interactions with surrounding objects cannot be explicitly simulated while out of view.
Spatial resolution limits control over small objects, while temporal resolution limits fast motion.
The model may also fail to follow physically implausible interventions, such as moving a solid object through a solid wall.

\section*{Acknowledgements}

This work was supported by the National Research Foundation of Korea (RS-2026-25497603: Fundamentals of Online 4D Reconstruction with Semantic Viewpoint Control and Generative Priors for Proactive Content Consumption); the IITP grant [RS-2020-II201361, Artificial Intelligence Graduate School Program (Yonsei University)] funded by the Korean government (MSIT); and the ``Advanced GPU Utilization Support Program'' funded by the Government of the Republic of Korea (Ministry of Science and ICT).

\bibliography{iclr2027_conference}
\bibliographystyle{iclr2027_conference}

\clearpage
\appendix
\section{Implementation details}

\subsection{UniState Construction}
\label{asec:gap_patch}

\paragraph{Geometry and feature alignment.}
Given an initial image $I_0$ and its camera $C_0$, we obtain latents $z_0=\mathcal{E}(I_0)$ using a VAE encoder and estimate depth $D_0$ using an off-the-shelf estimator~\citep{wang2026moge}.
Using $D_0$ and $C_0$, we associate each feature $z_{0,i}$ with 3D geometry $X_{0,i}$ in the world coordinate system defined by $C_0$.
These geometry--feature pairs form the initial state $S_0=\{\mathcal{P}_{0,i}\}_{i=1}^{N}$, where
$\mathcal{P}_{0,i}=(X_{0,i},z_{0,i})$ denotes a 3D latent.
Additional views of the same initial scene can extend $S_0$ by lifting and registering their latent features into this coordinate system.

Let \(\{\Omega_i\}_{i=1}^{N}\) denote the initial image patches. We construct
the geometric support of patch \(i\) as
\[
    X_{0,i}
    =
    \Pi^{-1}_{C_0}
    \left(
        \Omega_i,
        D_0[\Omega_i]
    \right),
\]
where \(\Pi^{-1}_{C_0}\) denotes unprojection through the source camera.
The implementation approximates the resulting local geometry with an
oriented planar primitive. Its appearance grid is sampled from the aligned
latent coordinates \(U_i\).

\paragraph{Intervention.}
Entity geometry is transformed according to
\cref{eq:entity_transform}.
At each time $t$, we construct
$S_t=\{\mathcal{P}_{t,i}\}_{i=1}^{N}$ directly from $S_0$,
where $\mathcal{P}_{t,i}=(X_{t,i},z_{0,i})$.
Thus, each transformed geometry remains paired with its
source latent feature.

\paragraph{Patch selection.}

Given an entity mask \(S_k\), patch \(i\) is assigned to entity \(k\) when
\[
    \frac{
        |\Omega_i\cap S_k|
    }{
        |\Omega_i|
    }
    >
    \tau_{\mathrm{seg}}.
\]
We store a control indicator
\(\alpha_i=\mathbb{1}[i\in\bigcup_k\mathcal{I}_k]\) separately from the
transform. This indicator remains active when the selected entity holds a
constant pose.

\paragraph{Back-face rejection.}

Let \(n_{i,t}\) and \(\bar{x}_{t,i}\) denote the transformed surface normal
and patch center. Given target camera center \(o_t\), patch \(i\) is rendered
when
\[
    \left\langle
        n_{i,t},
        o_t-\bar{x}_{t,i}
    \right\rangle
    >0.
\]
Patches at an angle of \(90^\circ\) or greater relative to the front-facing
viewing direction are rejected.

\paragraph{Intervention support.}

We first rasterize every selected and front-facing patch:
\[
    \widetilde{M}_t
    =
    \mathcal{R}_{\mathrm{support}}
    \left(
        \{
            X_{t,i}
        \}_{\alpha_i=1};
        C_t
    \right).
\]
A scene-level depth test then produces
\[
    M_t
    =
    \operatorname{Vis}
    \left(
        \widetilde{M}_t;
        \{X_{t,i}\}_{i=1}^{N},
        C_t
    \right).
\]
Thus, patches occluded by other scene geometry are absent from \(M_t\).

\paragraph{Adapter injection.}
The latent and intervention-mask conditions are injected at
the transformer input, whereas camera conditioning is applied
within each transformer block.

For the input conditions, let $h_{V,\mathrm{base}}^{(0)}$
denote the backbone's video-token embeddings after the input
projection and timestep embedding, before adding the $W$ and
$M$ conditions. We form the conditioned input tokens as
\[
    h_V^{(0)}
    =
    h_{V,\mathrm{base}}^{(0)}
    +
    P_W\!\left(\operatorname{patch}(W)\right)
    +
    P_M\!\left(\operatorname{patch}(M)\right).
\]
Here, $\operatorname{patch}$ arranges each condition into
patches aligned with the backbone's video-token grid.
$P_W$ and $P_M$ are separate trainable linear projections
without bias, mapping 192-dimensional latent patches and
4-dimensional binary mask patches, respectively, to the
4096-dimensional video-token space.
Each mask patch corresponds to a $2\times2$ spatial region.
These residuals are added once before the transformer blocks;
the $M$ term is omitted during the first training stage.

For camera conditioning, each transformer block contains
a trainable attention branch parallel to the backbone
self-attention. This branch computes queries, keys, and
values from the normalized video tokens and applies PRoPE
using the relative-ray representation derived from the
camera poses and intrinsics.
Let $a_{V,\mathrm{base}}^{(\ell)}$ denote the backbone
self-attention output for the video tokens at block $\ell$.
The camera-conditioned attention output is
\[
    a_V^{(\ell)}
    =
    a_{V,\mathrm{base}}^{(\ell)}
    +
    A_{\mathrm{cam}}^{(\ell)}
    \!\left(\operatorname{LN}(h_V^{(\ell)}),C\right),
\]
where $A_{\mathrm{cam}}^{(\ell)}$ denotes the parallel
camera-aware attention branch.
We apply this branch in all 36 transformer blocks with
a residual scale of 1.
The backbone remains frozen, while the camera branches
and the input-condition projections are trained according
to the two-stage curriculum.

\paragraph{Initialization from multiple observations.}
Although a single frame is sufficient to initialize the initial state,
patches from additional observations can be lifted into the same
world coordinate system to extend its spatial coverage.
The long sequence in \cref{fig:ours-qualitative}(b) uses an initial
state initialized from multiple observations and spans three
successive rollouts with object control and camera control.

\subsection{SAJG Construction}
\label{app:sajg}

\paragraph{Condition rendering.}
For each target camera $C_t$, the scene-anchor condition renders the initial state without interventions:
\begin{equation}
(W_{s,t},\_) = \mathcal{R}(S_0;C_t).
\end{equation}
The joint condition renders the intervened state:
\begin{equation}
(W_{j,t},M_t) = \mathcal{R}(S_t;C_t).
\end{equation}
Both conditions use the complete scene for depth testing.
The joint branch receives the intervention-mask condition, while the scene-anchor branch omits it.

\paragraph{Scene-mask construction.}
We construct $G_s$ in latent-video coordinates.
For each target camera, we project the selected entities at their source and target poses to obtain the binary support volumes $E_{\mathrm{src}}$ and $E_{\mathrm{tgt}}$.
The scene mask is
\begin{equation}
G_s
=
\mathcal{F}_{1,1}
\left[
1-
\mathcal{D}_{1}
\left(E_{\mathrm{src}}\cup E_{\mathrm{tgt}}\right)
\right],
\end{equation}
where $\mathcal{D}_{1}$ applies one-cell spatial dilation and $\mathcal{F}_{1,1}$ applies one-cell spatial and temporal feathering.
The resulting values lie in $[0,1]$, and the mask is broadcast across latent channels.
Unlike the intervention mask $M$, which marks the visible target entity regions, $G_s$ selects scene regions away from both source and target entity supports, with soft transitions at their boundaries.

The mask is geometry-derived rather than learned.
It may vary across video frames, but is computed only once for each input and remains unchanged across denoising steps.
All dilation and feathering settings are fixed; only $\alpha$ is varied in the routing ablation.

\paragraph{Equivalent correction form.}
Substituting \cref{eq:sajg_masked} into \cref{eq:sajg_final} gives
\begin{equation}
\begin{aligned}
V_{\mathrm{final}}
&=
V_j+(1-\alpha)G_s\odot(V_s-V_j)\\
&=
w_s\odot V_s+w_j\odot V_j,
\end{aligned}
\label{eq:sajg_correction}
\end{equation}
where
\begin{equation}
w_s=(1-\alpha)G_s,
\qquad
w_j=1-(1-\alpha)G_s.
\label{eq:sajg_weights}
\end{equation}
Thus, $w_s+w_j=1$ and $w_j\geq\alpha$ at every spatio-temporal location.
The equivalent correction form shows that the scene-anchor prediction modifies the joint prediction only where the scene mask permits.
At the default $\alpha=0.25$, the effective weights are $w_s=0.75G_s$ and $w_j=1-0.75G_s$.
Setting $\alpha=0$ applies the maximum scene correction permitted by the mask, while $\alpha=1$ removes the correction.

\paragraph{Classifier-free guidance and evaluation cost.}
Given the same noisy latent $x_\tau$ at flow-matching time $\tau$
and camera condition $C$, the conditional video model $f_\theta$
produces velocity predictions with classifier-free guidance (CFG):

\begin{equation}
V_s = \operatorname{CFG}
\bigl[f_\theta(x_\tau,\tau;C,W_s)\bigr], 
V_j = \operatorname{CFG}
\bigl[f_\theta(x_\tau,\tau;C,W_j)\bigr].
\label{eq:sajg_predictions}
\end{equation}

Both predictions share the source observation and text prompt;
these and the intervention-mask input are omitted from the notation.

The main-paper equations use predictions after classifier-free guidance.
In implementation, we apply the same spatial weights $w_s$ and $w_j$
separately to the conditional and null-text predictions, and then apply
classifier-free guidance to the two mixed predictions.
With the same guidance scale for both branches, this is algebraically
equivalent to mixing the guided predictions because both operations
are linear.
The exact computation uses a separate null-text prediction for each
branch and requires four transformer evaluations per denoising step.

We additionally tested an approximation that reuses a single null-text
prediction for both branches while retaining their separate conditional
predictions.
Although the two branch-specific null-text predictions are not
necessarily identical, we observed no noticeable differences in the
generated results under this approximation in the tested settings.
Reusing one null-text prediction reduces the computation to three
transformer evaluations per denoising step.

\subsection{Training}
\label{app:training}
\paragraph{Training settings and captions.}
Both stages use AdamW with a global batch size of 8
and a base learning rate of $1\times10^{-4}$.
We generate training-video captions using
Qwen2.5-VL~\citep{bai2025qwen25vl}.
The captions describe scene content while omitting motion
and object-location descriptions
(e.g., ``A is to the left of B'').

\paragraph{Optimization.}
Both stages use the flow-matching objective
\begin{equation}
\mathcal{L}_{\mathrm{FM}}
=
\mathbb{E}_{x,\tau}
\left[
\left\|
f_\theta(x_\tau,\tau;C,W,M)
-
V_\tau^{\mathrm{gt}}
\right\|_2^2
\right],
\end{equation}
with the $M$ condition omitted in the first stage.
The first stage trains the $C$ and $W$ conditioning modules on DL3DV
for 20K steps.
The second stage initializes these modules from the first-stage
checkpoint, adds the $M$ module, and jointly trains all three
conditioning paths on HM-World for 10K steps.
The video backbone, VAE, and text encoder remain frozen.

\paragraph{Training condition construction.}
We render $S_0$ from the target camera $C_t$ and zero out
the selected entities' projected source regions, obtaining
$W_t^{\mathrm{src}}$.
Let $z_t^e$ denote the ground-truth entity latent aligned with
the target latent grid, and $B_t$ its binary spatial support.
We form the training condition as
\begin{equation}
W_t^{\mathrm{train}}
=
(1-B_t)\odot W_t^{\mathrm{src}}
+
B_t\odot z_t^e,
\label{eq:training_compositing}
\end{equation}
with $B_t$ broadcast across latent channels.
This retains source-scene appearance outside the target entities,
while vacated source regions remain zero unless covered by
a target entity.

\paragraph{Conditioning augmentation.}
For a latent grid of size $H\times W$,
$\mathcal{D}_{\mathrm{NN}}$ reduces the spatial resolution to
$\operatorname{round}(\rho H)\times
\operatorname{round}(\rho W)$ using nearest-neighbor sampling.
$\mathcal{U}_{\mathrm{NN}}$ restores the original resolution using
nearest-neighbor upsampling.
Given the ground-truth entity latents $z_t^e$ and their binary
spatial support $B_t$, we define
\begin{equation}
\widetilde{z}_t^e =
\begin{cases}
z_0^e, & t=0,\\
(1-B_t)\odot z_t^e
+
B_t\odot
\mathcal{U}_{\mathrm{NN}}
\bigl(\mathcal{D}_{\mathrm{NN}}(z_t^e)\bigr),
& t>0.
\end{cases}
\label{eq:taa}
\end{equation}
As defined in \cref{eq:taa}, the resampled values replace the
conditioning appearance only within the future-frame entity support
$B_t$.
The support mask itself is not resampled.
The augmented entity latents are then composited with the initial-state
renders using \cref{eq:training_compositing}.
Neither the intervention mask $M$ nor the ground-truth training target
is modified.

\subsection{WCPE Evaluation}
\label{app:wcpe}

\paragraph{Motivation and target projections.}
An object's image-space position alone does not establish whether
the prescribed world-space control is realized relative to the scene.
WCPE therefore evaluates object--background relative positions
together with background alignment.
Target trajectories are constructed from the input scene proxy:
object points follow the prescribed object transformations,
while background points remain fixed in world coordinates.
Both are projected using the target camera trajectory.
These targets are shared across methods for each evaluation condition.

\paragraph{Metric formulation.}
We track object queries and background anchors in the generated
videos using CoTracker3~\citep{karaev2025cotracker3}.
For an object query at a given frame, let $\hat{p}$ and $p^*$
denote its tracked image position and target projection.
Similarly, $\hat{b}_j$ and $b_j^*$ denote the tracked and target
positions of background anchor $j$.
Using the same valid anchor set $\mathcal{V}$, we define
\begin{equation}
    e_{\mathrm{rel}}
    =
    \operatorname{median}_{j \in \mathcal{V}}
    \frac{
        \left\lVert
        (\hat{p}-\hat{b}_j)-(p^*-b_j^*)
        \right\rVert_2
    }{D},
    \label{eq:wcpe-relative}
\end{equation}
\begin{equation}
    e_{\mathrm{bg}}
    =
    \operatorname{median}_{j \in \mathcal{V}}
    \frac{
        \left\lVert \hat{b}_j-b_j^* \right\rVert_2
    }{D},
    \label{eq:wcpe-background}
\end{equation}
where $D=\sqrt{W^2+H^2}$ is the image diagonal.
The joint error is
\begin{equation}
    e_{\mathrm{joint}}
    =
    \max\left(e_{\mathrm{rel}},e_{\mathrm{bg}}\right).
    \label{eq:wcpe-joint}
\end{equation}
A low joint error requires both accurate object--background
relative positions and accurate background alignment with
the target projection.

\paragraph{Failure handling and aggregation.}
We report WCPE@5 with an error cap and missing-measurement
penalty of $\tau=0.05$, corresponding to 5\% of the image diagonal:
\begin{equation}
    e_{\mathrm{score}}
    =
    \begin{cases}
        \min(e_{\mathrm{joint}},\tau),
        & \text{if the measurement is valid},\\
        \tau,
        & \text{otherwise}.
    \end{cases}
    \label{eq:wcpe-capped}
\end{equation}
The eligible query--frame pairs are determined by target
visibility and finite target projections, independently of
the generated video.
Invalid measurements remain in this fixed evaluation population
and receive the maximum penalty.

Each video-level score is the mean of $e_{\mathrm{score}}$
over its eligible query--frame pairs.
We evaluate 277 scenarios with two independent random seeds
per scenario, yielding 554 videos per method.
The final WCPE is the equally weighted mean of all 554
video-level scores, reported in percentage units.
The same evaluation protocol is applied to all methods.

\paragraph{Interpretation.}
WCPE measures projection-level consistency with the prescribed
world-space object and camera controls.
It does not reconstruct or directly verify metric 3D trajectories,
and its accuracy depends on the scene proxy and point tracking.
\section{Additional ablations}
\label{app:ablations}

\subsection{Curriculum training}
\label{app:curriculum-ablation}

We compare the staged curriculum in \cref{sec:training} with
an interleaved baseline that mixes the same conditioning tasks
throughout training.
Both schedules use 30K total updates: 20K on DL3DV with $C+W$
conditioning and 10K on HM-World with $C+W+M$ conditioning.
All other training settings are identical, and we select the
best checkpoint for each schedule based on validation performance.

As shown in \cref{tab:ablation-curriculum}, staged training
achieves lower WCPE, Cam-T, and Cam-R.
Since both schedules use the same data exposure and update counts,
this comparison supports the benefit of training order.
We also continued the interleaved baseline beyond 30K steps
and observed no further validation improvement.

\begin{table}[t]
\centering
\caption{Staged versus interleaved training under matched
data exposure and computation.}
\label{tab:ablation-curriculum}
\begin{tabular}{lccc}
\toprule
Training configuration
& WCPE (\%) $\downarrow$
& Cam-T $\downarrow$
& Cam-R ($^\circ$) $\downarrow$ \\
\midrule
Interleaved training
& 2.630 & 0.0117 & 0.413 \\
Staged curriculum (Ours)
& \textbf{2.323} & \textbf{0.0106} & \textbf{0.376} \\
\bottomrule
\end{tabular}
\end{table}
\section{More comparisons}
\label{sec:more-comparisons}
In this section, we provide detailed representation, capability and qualitative comparison results. The green boxes and XYZ axes indicate the target object positions and orientations, respectively.

\begin{figure}[t]
\centering
\includegraphics[width=0.95\linewidth]{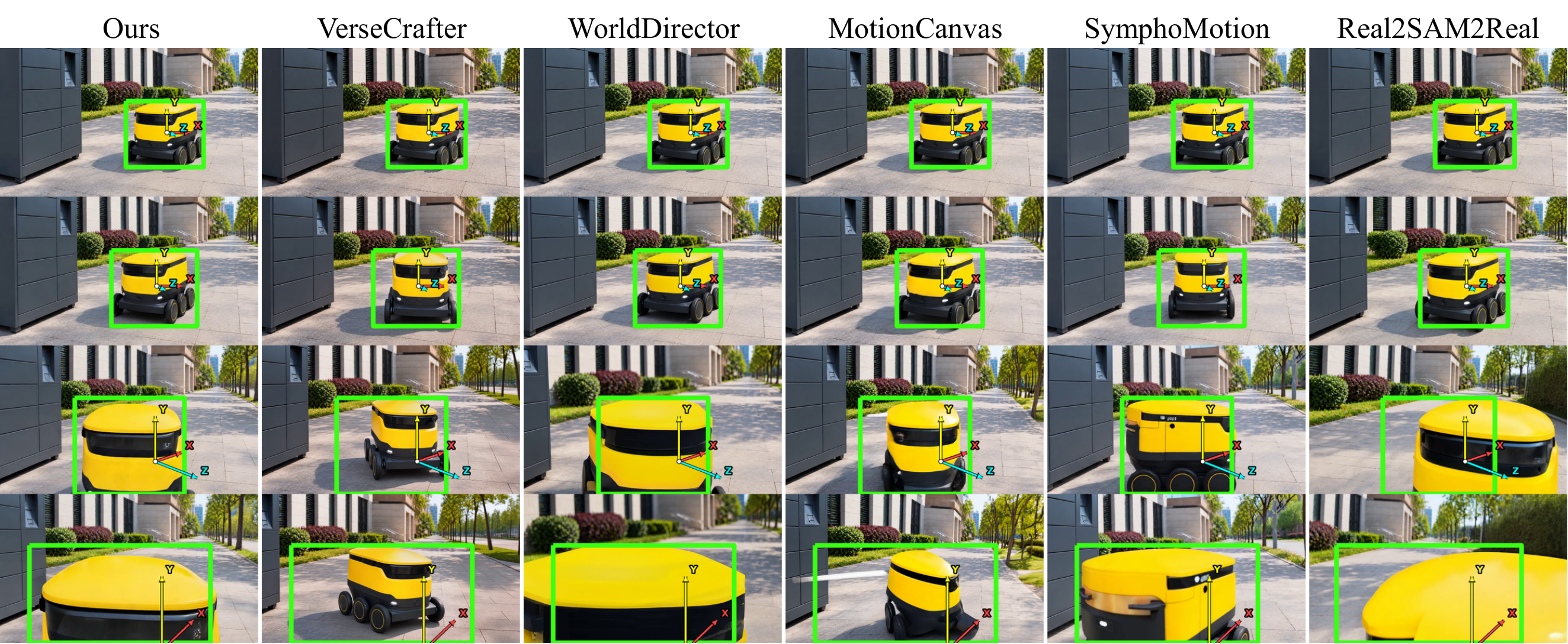}
\caption{\textbf{Qualitative comparison of object translation and rotation control}. The user translates the yellow robot forward and left while rotating it in yaw, and independently translates the camera to the right. Ours follows the robot's approach and turn under the prescribed viewpoint change. VerseCrafter shows insufficient forward motion; MotionCanvas and SymphoMotion deviate from the target heading; and WorldDirector and Real2SAM2Real produce excessive enlargement accompanied by unintended scene motion.}
\label{fig:vs-competitors-deliverrobots}
\end{figure}

\begin{figure}[t]
\centering
\includegraphics[width=0.95\linewidth]{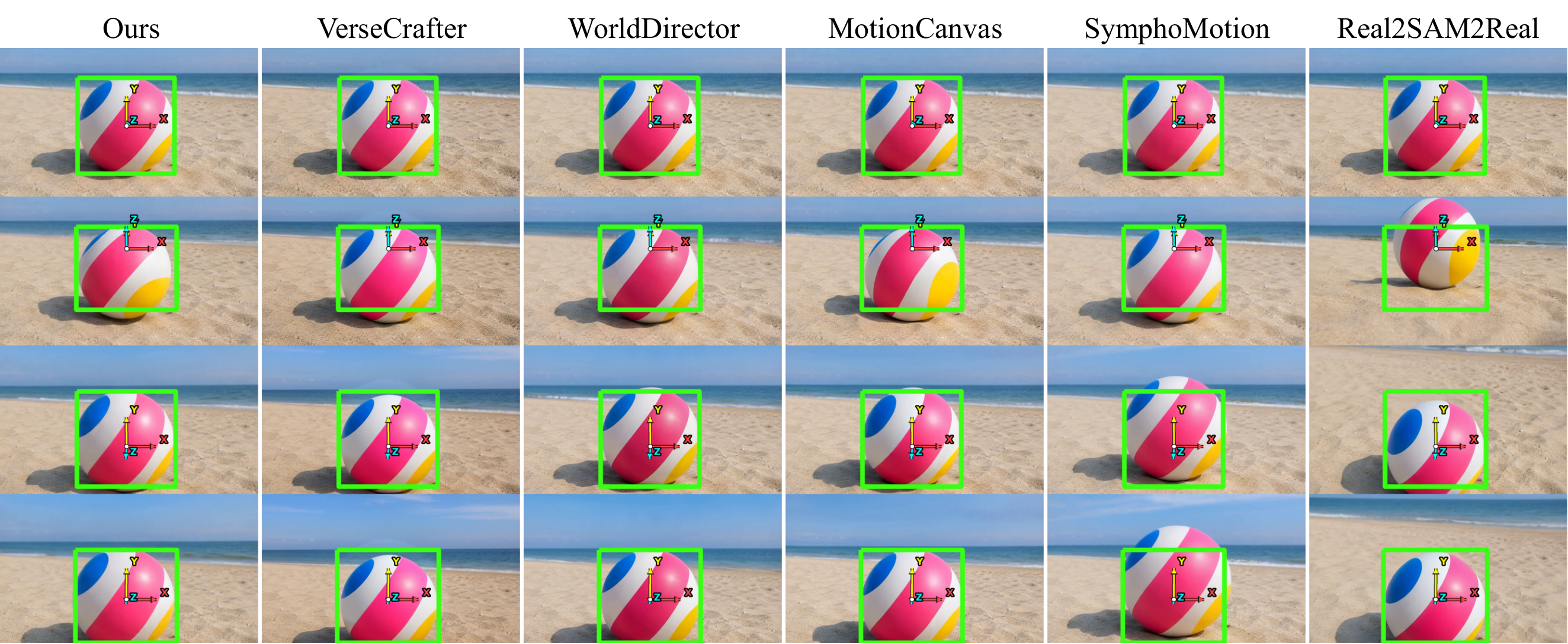}
\caption{\textbf{Qualitative comparison of object rotation control}. The user specifies vertical oscillation and pitch rotation for the ball while pitching the camera upward. VerseCrafter, WorldDirector, SymphoMotion, and Real2SAM2Real fail to reflect the prescribed ball rotation. Ours reproduces the ball's changing orientation and vertical motion under the moving viewpoint.}
\label{fig:vs-competitors-beachball}
\end{figure}

\begin{figure}[t]
\centering
\includegraphics[width=0.95\linewidth]{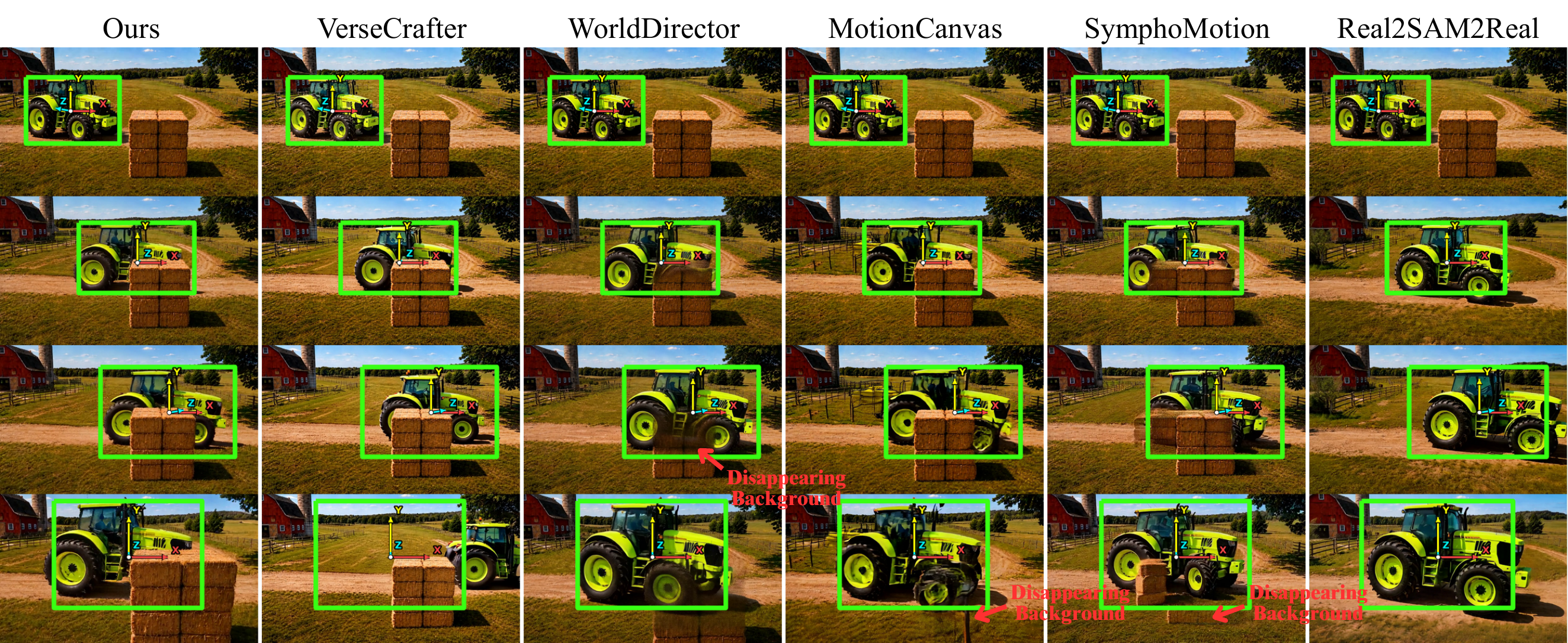}
\caption{\textbf{Qualitative comparison of occlusion handling}. The user moves the tractor forward behind the hay bales and then partially back while dollying the camera forward. WorldDirector, MotionCanvas, and SymphoMotion cause the hay bales to disappear. Real2SAM2Real absorbs the tractor's intended motion into unintended viewpoint change, while VerseCrafter fails to reproduce its partial return. Ours follows both the tractor trajectory and camera dolly while preserving the bales and their occlusion of the tractor.}
\label{fig:physical-registry}
\end{figure}

\begin{table*}[t]
    \centering
    \caption{
    Capability comparison of object-control methods.
    Appearance-bearing scene memory denotes spatially grounded appearance
    evidence that can be reused across viewpoints.
    Explicit 3D object control indicates that object position or pose can be
    directly specified in 3D.
    Occlusion-aware appearance transport requires transporting appearance
    evidence with explicit depth and visibility reasoning.
    \cmark, \tmark, and \xmark denote full, partial or indirect, and unsupported
    capabilities, respectively.
    }
    \label{tab:capability_comparison}
    \begin{tabular}{lccc}
    \toprule
    Method
    & \makecell{Appearance-bearing\\scene memory}
    & \makecell{Explicit 3D\\object control}
    & \makecell{Occlusion-aware\\appearance transport} \\
    \midrule
    VerseCrafter
        & \cmark & \cmark & \tmark \\
    WorldDirector
        & \xmark & \tmark & \xmark \\
    MotionCanvas
        & \xmark & \tmark & \xmark \\
    SymphoMotion
        & \tmark & \cmark & \tmark \\
    Real2SAM2Real
        & \xmark & \cmark & \xmark \\
    \textbf{Ours}
        & \cmark & \cmark & \cmark \\
    \bottomrule
    \end{tabular}

\end{table*}

\begin{table*}[t]
    \centering
    \caption{
    Detailed comparison of representations used for user intervention.
    Geometry denotes the representation used to instantiate the intervention,
    while appearance indicates whether the representation explicitly carries appearance information.
    A dash indicates no explicit appearance information.
    }
    \label{tab:intervention_representation}
    \begin{tabular}{lll}
    \toprule
    \multirow{2}{*}{Method}
    & \multicolumn{2}{c}{User intervention} \\
    \cmidrule(lr){2-3}
    & \multicolumn{1}{c}{Geometry}
    & \multicolumn{1}{c}{Appearance} \\
    \midrule
    VerseCrafter
        & 3D Gaussian trajectories
        & -- \\
    WorldDirector
        & 2D box trajectories from 3D plans
        & Box-aligned RGB crops \\
    MotionCanvas
        & 2D boxes / point tracks
        & -- \\
    SymphoMotion
        & 3D point trajectories
        & -- \\
    Real2SAM2Real
        & 3D mesh cache $\rightarrow$ normal maps
        & -- \\
    \textbf{Ours}
        & \textbf{Transformed 3D patches}
        & \textbf{Latent appearance grids} \\
    \bottomrule
    \end{tabular}

\end{table*}

\begin{table}[t]
\caption{Comparison with MosaicMem under the same evaluation conditions as UniState. MosaicMem supports camera control but does not natively support object control. We apply time-varying 6-DoF transformations to selected object patches for comparison. Under these interventions, we observe diverse failure modes, including camera-control failures and objects remaining stationary or disappearing instead of following the specified motion, as illustrated in \cref{fig:mosaicmem_failure_qual}.}
\label{app:comparison-mosaicmem}
\begin{center}
\begin{tabular}{lccc}
\toprule
Method & WCPE (\%) $\downarrow$ & Cam-T $\downarrow$ & Cam-R ($^\circ$) $\downarrow$
\\ 
\midrule
MosaicMem (reproduced) & 2.522 & 0.01249 & 0.5132 \\
Ours & \textbf{2.3233} & \textbf{0.0106} & \textbf{0.3761} \\
\bottomrule
\end{tabular}
\end{center}
\end{table}

\begin{figure}[t]
\centering
\includegraphics[width=0.85\linewidth]{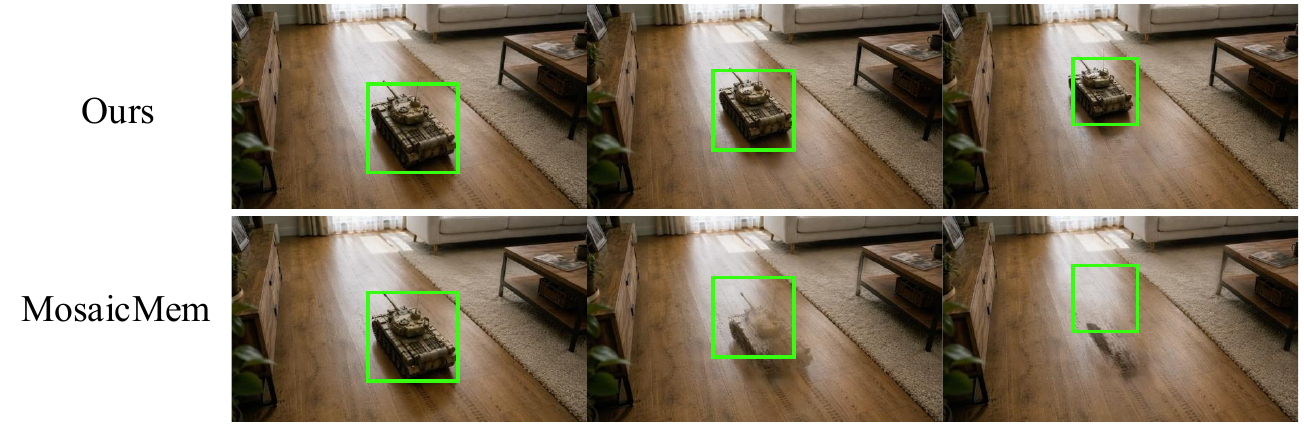}
\caption{Qualitative failure of MosaicMem in an object-only intervention
scenario with a fixed camera. The tank fails to follow the specified
motion and disappears midway through the generated sequence.}
\label{fig:mosaicmem_failure_qual}
\end{figure}

\begin{figure}[t]
\centering
\includegraphics[width=0.95\linewidth]{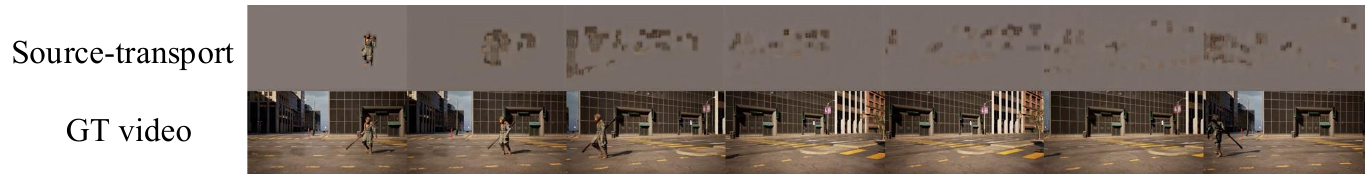}
\caption{Misalignment in source-transport conditions. Rows show the source-transport condition (top) and ground-truth video (bottom) for the same sequence. Source transport produces floating artifacts due to inaccurate geometric alignment. TAA instead augments ground-truth entity latents to introduce transport-like appearance distortions while preserving the target entity support, avoiding the misaligned floaters present in the source-transport condition.}
\label{fig:source-transport-visualization}
\end{figure}

\end{document}